# Artist Agent: A Reinforcement Learning Approach to Automatic Stroke Generation in Oriental Ink Painting

**Ning Xie**  XIE@SG.CS.TITECH.AC.JP
**Hirotaka Hachiya**  HACHIYA@SG.CS.TITECH.AC.JP
**Masashi Sugiyama**  SUGI@CS.TITECH.AC.JP
Department of Computer Science, Tokyo Institute of Technology, Tokyo, Japan

## Abstract

Oriental ink painting, called *Sumi-e*, is one of the most appealing painting styles that has attracted artists around the world. Major challenges in computer-based Sumi-e simulation are to abstract complex scene information and draw smooth and natural brush strokes. To automatically generate such strokes, we propose to model a brush as a reinforcement learning agent, and learn desired brush-trajectories by maximizing the sum of rewards in the policy search framework. We also elaborate on the design of actions, states, and rewards tailored for a Sumi-e agent. The effectiveness of our proposed approach is demonstrated through simulated Sumi-e experiments.

## 1. Introduction

Among various techniques of non-photorealistic rendering, stroke-based painterly rendering simulates common practices of human painters who create paintings with brush strokes. In this paper, we focus on oriental ink painting.

Unlike western styles, such as water-color, pastel, and oil painting, which place overlapped strokes into multiple layers (Hertzmann, 1998; Shiraishi & Yamaguchi, 2000), oriental ink painting uses few strokes to convey significant information about the scene. An artist can draw expressive strokes in various styles by soft brush tufts. The appearance of the stroke is therefore determined by the shape of an object to paint, the path and posture of the brush, and the distribution of pigments in the brush.

Drawing smooth and natural strokes in arbitrary shapes is challenging since an optimal brush trajectory and the posture of a brush footprint[1] are different for each shape. Xie et al. (2011) formulated the problem of drawing brush strokes as minimization of an accumulated energy of moving the brush and used the Dynamic Programming (DP) approach to obtain optimal brush strokes. It was demonstrated that smooth and natural brush strokes could be obtained by minimizing the accumulated energy. However, the stroke optimized by DP for a specific shape cannot be applied to other shapes even when the difference is small. Thus, it is not efficient if the target object is composed of many basic shapes, e.g., a Chinese character, since the optimal brush stroke for each shape has to be obtained. Furthermore, ordinary DP cannot directly handle continuous actions and states. Thus, smoothness of resulted brush strokes is highly dependent on the discretization of spaces.

In this paper, we introduce a reinforcement learning (RL) approach to solving this problem. We model a soft-tuft brush as an RL agent that makes a sequential decision on which direction to move, and train the agent to draw graceful strokes in arbitrary shapes (see Figure 1). Our idea is to first learn a desired drawing policy by maximizing the sum of rewards from a number of typical training shapes. Then, the trained policy is applied to draw strokes in various new shapes.

More specifically, the proposed approach contains two technical challenges: how to design the brush agent and how to train the agent's policy. We first propose to design the state space of the brush agent to be relative to its surrounding shape, e.g., boundaries and the medial axis, to learn a general drawing policy which is independent of a specific entire shape. Secondly, we propose to formulate stroke drawing by a Markov decision process (MDP) (Sutton & Barto, 1998) and apply a *policy gradient* method (Williams, 1992) to learn a (local) optimal drawing policy. An advantage of the policy gradient method is that it can naturally handle continuous states and actions which are important for obtaining smooth and natural brush strokes. Furthermore, since a policy is a function for selecting actions given a state, a learned policy can be naturally applied to new shapes.

---

[1] We use a *footprint* to denote the region of a canvas which a brush stamps on.

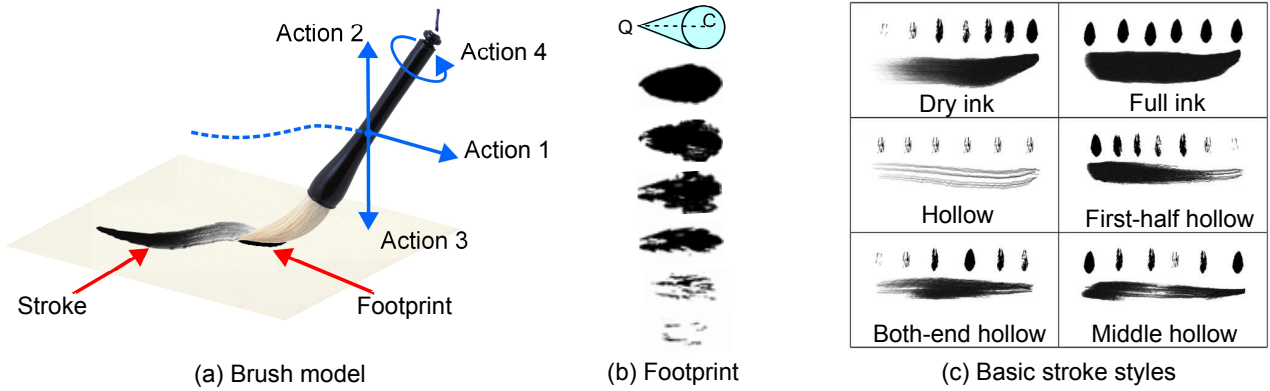

*Figure 1.* Illustration of our brush agent and its path. (a) In our model, a stroke is generated by moving the brush with the following 4 actions: Action 1 is the brush movement along the current path, Action 2 is pushing down the brush, Action 3 is lifting up the brush, and Action 4 is rotation of the brush handle. (b) The top symbol illustrates our brush agent, which consists of a tip $Q$ and a circle with center $C$ and radius $r$. Others illustrate footprints of a real brush with different ink quantities. (c) There are 6 basic stroke styles: dry ink, full ink, hollow, first-half hollow, both-end hollow, and middle hollow. The small footprints on the top of each stroke show the interpolation order.

## 2. Formulation of Automatic Stroke Generation

In this section, we formulate the problem of automatic stroke generation as a reinforcement learning (RL) problem.

### 2.1. Markov Decision Process

Let us formulate the procedure of drawing a stroke as a Markov Decision Process (MDP) consisting of a tuple $(\mathcal{S}, \mathcal{A}, p_\mathrm{I}, p_\mathrm{T}, R)$, where $\mathcal{S}$ is a set of continuous states, $\mathcal{A}$ is a set of continuous actions, $p_\mathrm{I}$ is the probability-density of the initial state, $p_\mathrm{T}(\boldsymbol{s}'|\boldsymbol{s},a)$ is the transition probability-density from the current state $\boldsymbol{s} \in \mathcal{S}$ to next state $\boldsymbol{s}' \in \mathcal{S}$ when taking action $a \in \mathcal{A}$, $R(\boldsymbol{s}, a, \boldsymbol{s}')$ is an immediate reward function for the transition from $\boldsymbol{s}$ to $\boldsymbol{s}'$ by taking action $a$.

Let $\pi(a|\boldsymbol{s};\boldsymbol{\theta})$ be a stochastic policy with parameter $\boldsymbol{\theta}$, which represents the conditional probability density of taking action $a$ given state $\boldsymbol{s}$. Let $h = (\boldsymbol{s}_1, a_1, \ldots, \boldsymbol{s}_T, a_T, \boldsymbol{s}_{T+1})$ be a trajectory of length $T$. Then the *return* (i.e., the discounted sum of future rewards) along $h$ is expressed as

$$R(h) = \sum_{t=1}^{T} \gamma^{t-1} R(\boldsymbol{s}_t, a_t, \boldsymbol{s}_{t+1}),$$

where $\gamma \in [0, 1)$ is the discount factor for the future reward.

The expected return for parameter $\boldsymbol{\theta}$ is defined by

$$J(\boldsymbol{\theta}) = \int p(h|\boldsymbol{\theta}) R(h) \mathrm{d}h,$$

where

$$p(h|\boldsymbol{\theta}) = p(\boldsymbol{s}_1) \prod_{t=1}^{T} p(\boldsymbol{s}_{t+1}|\boldsymbol{s}_t, a_t) \pi(a_t|\boldsymbol{s}_t, \boldsymbol{\theta}).$$

The goal of RL is to find the optimal policy parameter $\boldsymbol{\theta}^*$ that maximizes the expected return $J(\boldsymbol{\theta})$:

$$\boldsymbol{\theta}^* \equiv \mathrm{argmax}\ J(\boldsymbol{\theta}).$$

### 2.2. Policy Gradient Method

We use a *policy gradient* algorithm (Williams, 1992) to solve the above RL problem. That is, the policy parameter $\boldsymbol{\theta}$ is updated via gradient ascent as

$$\boldsymbol{\theta} \longleftarrow \boldsymbol{\theta} + \varepsilon \nabla_{\boldsymbol{\theta}} J(\boldsymbol{\theta}),$$

where $\varepsilon$ is a learning rate. The gradient $\nabla_{\boldsymbol{\theta}} J(\boldsymbol{\theta})$ is given by

$$\begin{aligned}\nabla_{\boldsymbol{\theta}} J(\boldsymbol{\theta}) &= \int \nabla_{\boldsymbol{\theta}} p(h|\boldsymbol{\theta}) R(h) \mathrm{d}h \\ &= \int p(h|\boldsymbol{\theta}) \nabla_{\boldsymbol{\theta}} \log p(h|\boldsymbol{\theta}) R(h) \mathrm{d}h \\ &= \int p(h|\boldsymbol{\theta}) \sum_{t=1}^{T} \nabla_{\boldsymbol{\theta}} \log \pi(a_t|\boldsymbol{s_t}, \boldsymbol{\theta}) R(h) \mathrm{d}h,\end{aligned}$$

where we used the so-called *log trick*: $\nabla_{\boldsymbol{\theta}} p(h|\boldsymbol{\theta}) = p(h|\boldsymbol{\theta}) \nabla_{\boldsymbol{\theta}} \log p(h|\boldsymbol{\theta})$. Since $p(h|\boldsymbol{\theta})$ is unknown, the expectation is approximated by the empirical average:

$$\nabla_{\boldsymbol{\theta}} \widehat{J}(\boldsymbol{\theta}) = \frac{1}{N} \sum_{n=1}^{N} \sum_{t=1}^{T} \nabla_{\boldsymbol{\theta}} \log \pi(a_t^{(n)}|\boldsymbol{s}_t^{(n)}, \boldsymbol{\theta}) R(h^{(n)}),$$

where $\{h^{(n)}\}_{n=1}^{N}$ are $N$ episodic samples with $T$ steps and $h^{(n)} = (s_1^{(n)}, a_1^{(n)}, \ldots, s_T^{(n)}, a_T^{(n)}, s_{T+1}^{(n)})$.

Let us employ the Gaussian policy function with parameter $\boldsymbol{\theta} = (\boldsymbol{\mu}^\top, \sigma)^\top$, where $\boldsymbol{\mu}$ is the mean vector and $\sigma$ is the standard deviation:

$$\pi(a|\boldsymbol{s}; \boldsymbol{\theta}) = \frac{1}{\sigma\sqrt{2\pi}} \exp\left(-\frac{(a - \boldsymbol{\mu}^\top \boldsymbol{s})^2}{2\sigma^2}\right).$$

Then the derivatives of the expected return $J(\boldsymbol{\theta})$ with respect to the parameter $\boldsymbol{\theta}$ are given as

$$\nabla_{\boldsymbol{\mu}} \log \pi(a|\boldsymbol{s}; \boldsymbol{\theta}) = \frac{a - \boldsymbol{\mu}^\top \boldsymbol{s}}{\sigma^2} \boldsymbol{s},$$

$$\nabla_\sigma \log \pi(a|\boldsymbol{s}; \boldsymbol{\theta}) = \frac{(a - \boldsymbol{\mu}^\top \boldsymbol{s})^2 - \sigma^2}{\sigma^3}.$$

Consequently, the policy gradients $\nabla_{\boldsymbol{\theta}} \widehat{J}(\boldsymbol{\theta})$ are expressed as

$$\nabla_{\boldsymbol{\mu}} J(\boldsymbol{\theta}) = \frac{1}{N} \sum_{n=1}^{N} (R(h^{(n)}) - b) \sum_{t=1}^{T} \frac{(a_t^{(n)} - \boldsymbol{\mu}^\top \boldsymbol{s}_t^{(n)}) \boldsymbol{s}_t^{(n)}}{\sigma^2},$$

$$\nabla_\sigma J(\boldsymbol{\theta}) = \frac{1}{N} \sum_{n=1}^{N} (R(h^{(n)}) - b) \sum_{t=1}^{T} \frac{\left(a_t^{(n)} - \boldsymbol{\mu}^\top \boldsymbol{s}_t^{(n)}\right)^2 - \sigma^2}{\sigma^3},$$

where $b$ is a baseline for reducing the variance of gradient estimates. The optimal baseline that minimizes the variance of the gradient estimate is given as follows (Peters & Schaal, 2006):

$$b^* = \underset{b}{\operatorname{argmin}} \mathbf{Var}[\nabla_\theta \widehat{J}(\theta)]$$

$$\simeq \frac{\frac{1}{N} \sum_{n=1}^{N} R(h^{(n)}) \left\| \sum_{t=1}^{T} \nabla_\theta \log \pi(a_t^{(n)}|\boldsymbol{s}_t^{(n)}; \boldsymbol{\theta}) \right\|^2}{\frac{1}{N} \sum_{n=1}^{N} \left\| \sum_{t=1}^{T} \nabla_\theta \log \pi(a_t^{(n)}|\boldsymbol{s}_t^{(n)}; \boldsymbol{\theta}) \right\|^2}.$$

Finally, the policy parameter $\boldsymbol{\theta} = (\boldsymbol{\mu}^\top, \sigma)^\top$ is updated as

$$\boldsymbol{\mu} \leftarrow \boldsymbol{\mu} + \varepsilon \nabla_{\boldsymbol{\mu}} J(\boldsymbol{\theta}),$$

$$\sigma \leftarrow \sigma + \varepsilon \nabla_\sigma J(\boldsymbol{\theta}).$$

## 3. Design of MDP for Brush Agent

In this section, we give a specific design of state space $\mathcal{S}$, action space $\mathcal{A}$, and immediate reward function $R(\boldsymbol{s}, a, \boldsymbol{s}')$ for our brush agent to learn smooth and natural strokes. For this purpose, we first extract the boundary of a given object and then calculate the medial axis $M$, as illustrated in Figure 2.

### 3.1. Design of Actions

To generate elegant brush strokes, the brush agent should move inside the given boundaries properly. To this end, we consider four basic actions of the brush agent: movement of the brush, scaling up/down of the footprint, and rotation of the heading direction of the brush (see Figure 1(a)).

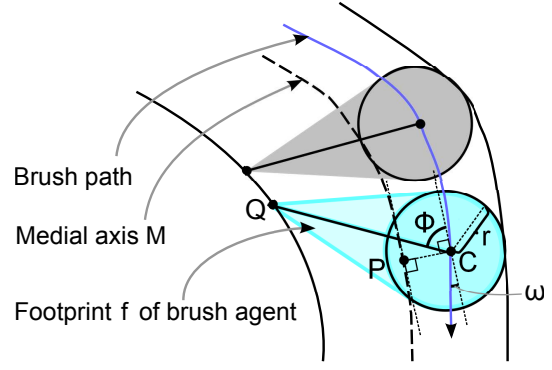

Figure 2. Illustration of brush agent and its path. The brush agent consists of a tip $Q$ and a circle with center $C$ and radius $r$. $P$ is the nearest point on $M$ to $C$.

Since properly covering the whole desired region is the most important issue, we treat the movement of the brush as the primary action (Action 1). The action $a$ specifies the angle of the velocity vector of the agent relative to the medial axis. The action is determined by the Gaussian policy function. In practical applications, the agent should be able to deal with arbitrary strokes in various scales. To achieve stable performance in different scales, we adaptively change the velocity of the brush movement relative to the scale of the current footprint. The other actions (Actions 2, 3, and 4) are automatically optimized to satisfy the assumption that the tip of the agent should touch one side of the boundary; meanwhile, the bottom of the agent should tangent with the other side of the boundary. Otherwise, a new footprint will remain the same posture as the previous one, but just transit to a new position by Action 1.

### 3.2. Design of States

We use the global measurement (the pose configuration of a footprint under the global Cartesian coordinate) and the relative state (the brush agent's pose and the locomotion information relative to the local surrounding environment). The relative state is calculated based on the global measurement values. Thus, both the global measurement and the relative state should be regarded as a state in terms of an MDP. However, for the calculating return and a policy, we use only the relative state, which allows the agent to learn a drawing policy that can be generalized to new shapes.

Our relative state space design consists of two parts: Current surrounding shape and upcoming shape. More specifically, our state space is expressed by six features $\boldsymbol{s} = (\omega, \phi, d, \kappa_1, \kappa_2, l)^\top$ (see Figures 2 and 3), where

- $\omega \in (-\pi, \pi]$: The angle of the velocity vector of the brush agent relative to the medial axis.

- $\phi \in (-\pi, \pi]$: The heading direction of the brush agent relative to the medial axis.

- $d \in [-2, 2]$: The ratio of offset distance $\delta$ (see Figure 3) from the center $C$ of the brush agent to the nearest point $P$ on the medial axis $M$ over the radius $r$ of the brush agent ($|d| = \delta/r$). $d$ takes positive/negative values when the center of the brush agent is on the left-/right-hand side of the medial axis. On the other hand, $d$ takes the value $0$ when the center of the brush agent is on the medial axis. Furthermore, when $d$ takes a value in $[-1, 1]$, the brush agent is inside the boundaries (for example, $d_{t-1}$ in Figure 3), and when the value of $d$ is in $[-2, -1)$ or in $(1, 2]$, the brush agent goes over the boundary of one side (for example, $d_t$ in Figure 3). In our system, the center of the agent is restricted within the shape. Therefore, the extreme value of $d$ is $\pm 2$, which means that the center of the agent is on the boundary.

- $\kappa_i (i = 1, 2) \in [0, 1)$: $\kappa_1$ provides the current surrounding information on the point $P_t$, whereas $\kappa_2$ provides the upcoming shape information on point $P_{t+1}$, as illustrated in Figure 3. The values are calculated as

$$|\kappa_i| = \frac{2}{\pi} \arctan\left(\frac{\alpha}{\sqrt{r'_i}}\right),$$

where $\alpha$ is the parameter to control the sensitivity to the curvature and $r'_i$ is the radius of the curve. More specifically, the value takes 0/negative/positive when the shape is straight/left-curved/right-curved, and the larger the value is, the tighter the curve is. Throughout this paper, we use a fixed value $\alpha = 0.05$.

- $l \in \{0, 1\}$: A binary label that indicates whether the agent moves to a region covered by the previous footprints or not. $l = 0$ means that the agent moves to a region covered by the previous footprint. Otherwise, $l = 1$ means that it moves to an uncovered region.

### 3.3. Design of Immediate Rewards

We design the reward function so that the smoother the brush stroke is, the higher the reward is. For this purpose, we define the reward function as

$$R(s_t, a_t, s_{t+1}) = \begin{cases} 0 & \text{if } f_t = f_{t+1} \text{ or } l = 0, \\ \dfrac{1 + (|\kappa_1(t)| + |\kappa_2(t)|)/2}{\lambda_1 E_{\text{location}}^{(t)} + \lambda_2 E_{\text{posture}}^{(t)}} & \text{otherwise.} \end{cases}$$

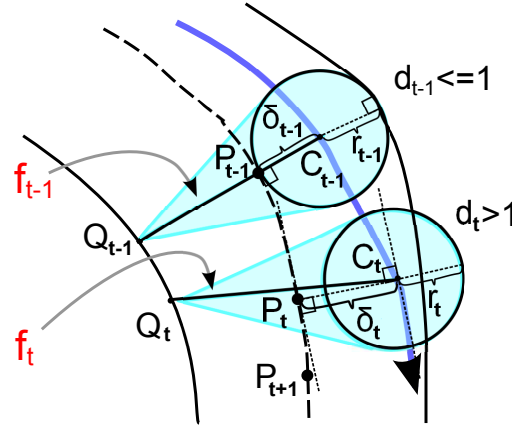

Figure 3. Illustration of the ratio $d$ of the offset distance $\delta$ over the radius $r$. Footprint $f_{t-1}$ is inside the drawing area. The circle with center $C_{t-1}$ and the tip $Q_{t-1}$ touch the boundary on each side. In this case, $\delta_{t-1} \leq r_{t-1}$ and $d_{t-1} \in [0, 1]$. On the other hand, $f_t$ goes over the boundary, and $\delta_t > r_t$ and $d_t > 1$. In our implementation, we restrict $d$ to be in $[-2, 2]$.

That is, the immediate reward is zero when the brush is blocked by a boundary as $f_t = f_{t+1}$ or the brush is going backward to a region that has already been covered by previous footprints $f_i$ ($i < t+1$). $|\kappa_1(t)| + |\kappa_2(t)|$ adaptively increases immediate rewards depending on the difficulty of the current shape measured by the curvature $\kappa_i(t)$ of the medial axis.

$E_{\text{location}}^{(t)}$ measures the quality of the location of the brush agent with respect to the medial axis, defined by

$$E_{\text{location}}^{(t)} = \begin{cases} \tau_1 |\omega_t| + \tau_2(|d_t| + W) & d \in [-2, -1) \cup (1, 2], \\ \tau_1 |\omega_t| & d \in [-1, 1], \end{cases}$$

where $\tau_1$ and $\tau_2$ are weight parameters and $W$ is the penalty. Since $d$ contains information whether the agent goes over the boundary or not, as illustrated in Figure 3, the penalty $W$ is added to $E_{\text{location}}$ when the agent goes over the boundary of the shape. When the brush agent is inside of the boundary, i.e., $d \in [-1, 1]$, $E_{\text{location}}$ depends only on the angle $\omega_t$ of the velocity vector.

$E_{\text{posture}}^{(t)}$ measures the quality of the posture of the brush agent based on neighboring footprints, defined by

$$E_{\text{posture}}^{(t)} = \zeta_1 \Delta \omega_t + \zeta_2 \Delta \phi_t + \zeta_3 \Delta d_t,$$

where $\Delta \omega_t$, $\Delta \phi_t$, and $\Delta d_t$ are changes in angles $\omega$ of the velocity vector, heading directions $\phi$, and ratios $d$ of the offset distance, respectively. The notation $\Delta x_t$ denotes the normalized squared changes between $x_{t-1}$ and $x_t$ defined by

$$\Delta x_t = \begin{cases} 1 & \text{if } x_t = x_{t-1} = 0, \\ \dfrac{(x_t - x_{t-1})^2}{(|x_t| + |x_{t-1}|)^2} & \text{otherwise.} \end{cases}$$

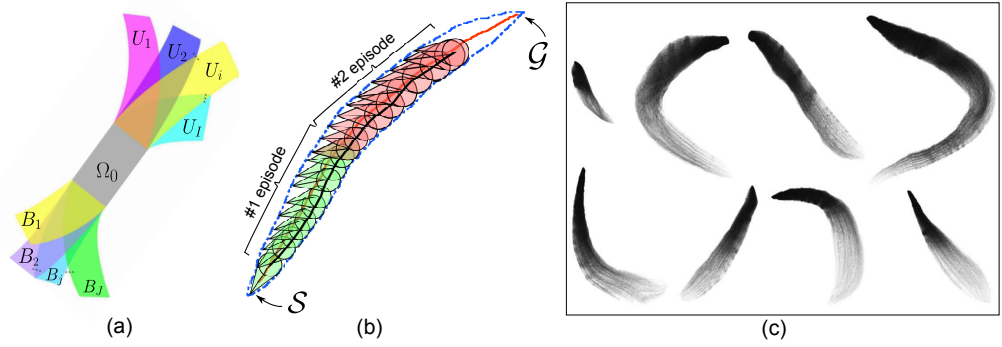

*Figure 4.* Policy training scheme. (a) Shape combination. Each shape $(U_i \cup \Omega_0 \cup L_j, i = 1, 2, ..., I$ and $j = 1, 2, ..., J)$ is combined with one of the upper regions $U_i$, the common region $\Omega_0$, and the lower regions $L_j$. (b) Setup of policy training. $\mathcal{S}$ is the start point of the shape. $\mathcal{G}$ is the goal point of the shape. The direction of the shape is from $\mathcal{S}$ to $\mathcal{G}$. (c) The brush library of single strokes in typical shapes. Only 8 out of 80 are shown here.

$\zeta_1, \zeta_2$, and $\zeta_3$ are weight parameters. We set the parameters at $\lambda_1 = \lambda_2 = 0.5, \tau_1 = \tau_2 = 0.5$, and $\zeta_1 = \zeta_2 = \zeta_3 = 1/3$.

### 3.4. Design of Training Sessions for Brush Agent

Given an appropriately designed MDP, the final step is to design training sessions, which is also highly important to make the brush agent useful in practice.

First of all, we propose to train the agent based on partial shapes, not the entire shapes. An advantage of this local training strategy is that various local shapes can be generated from a single entire shape, which significantly increases the number and variation of training samples. Another merit is that the generalization ability to new shapes can be enhanced, because even when the entire profile of a new shape is quite different from that of training data, the new shape may contain similar local shapes as illustrated in Figure 4(a).

To provide a wide variety of local shapes to the agent as training data, we prepared an in-house stroke library. This library contains 80 digitized real single brush strokes that are commonly used in Oriental ink painting. See Figure 4(c) for some examples. We extracted boundaries as the shape information and arranged them in a queue as training samples (see Figure 4(b)).

In the training scheme, the initial position of the first episode is chosen to be the start point $\mathcal{S}$ of the medial axis (Chin et al., 1995), and the direction to move is chosen to be the goal point $\mathcal{G}$ as illustrated in Figure 4(b). We estimate the length of an episode, $T$ from the selected shapes. In the first episode, the initial footprint is set around the start point of the shape. In the following episodes, the initial footprint is set as either the last footprint in the previous episode or the footprint around the start point. It depends on whether the agent moves well or is blocked by the boundary. For each policy, we repeat $N$ episodes to collect data $H = [h^{(1)}, h^{(2)}, \ldots, h^{(N)}]$, where $h^{(n)} = [\boldsymbol{s}_1^{(n)}, a_1^{(n)}, \ldots, \boldsymbol{s}_T^{(n)}, a_T^{(n)}, \boldsymbol{s}_{T+1}^{(n)}]$. We then use the data $H$ to calculate the gradient of the return, $\nabla_{\boldsymbol{\theta}} J(\boldsymbol{\theta})$, and update the policy parameter $M$ times to optimize the policy.

To ensure the continuity along the episodes, we design the initial location of the agent as shown in Figure 4(b): In the first episode, the initial location of the agent is set on the medial axis, with its tip $Q$ pointing to the end corner. In the next episode, the initial location is set to the location of the last footprint in the previous episode.

There are two exceptional situations where the new episode's initial location goes back to the initial location of the previous episode: The first situation is that the length of the up-coming track is much less than the length of the trajectory in $T$ steps. The other situation is that the agent is blocked by the footprints generated by the previous episode.

## 4. Experiments

In this section, we report experimental results.

### 4.1. Setup

We train the policy of the brush agent on the shape shown in Figure 4(c) through our training strategy introduced in Section 3.4. The parameter of the initial policy is set as $\boldsymbol{\theta} = (\boldsymbol{\mu}^\top, \sigma)^\top = (0, 0, 0, 0, 0, 0, 2)^\top$ according to the previous domain knowledge. The agent collects $N = 300$ episodic samples with trajectory length $T = 32$. The discount factor is set to $\gamma = 0.99$. The learning rate $\varepsilon$ is set as $0.1/\|\nabla_{\boldsymbol{\theta}} J_{\boldsymbol{\theta}}\|$. We investigate the average return over 10 trials as functions of policy-update iterations. The return at

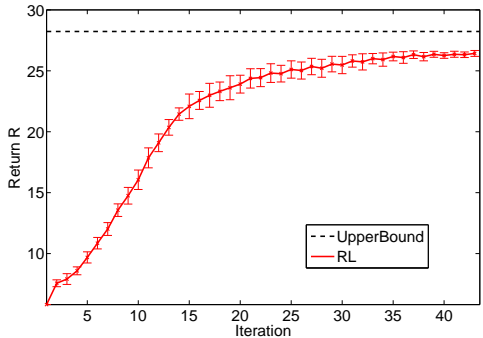

Figure 5. Average return over 10 trials for the RL method and the upper limit of the return value. The error bars denotes the standard deviation over 10 trials.

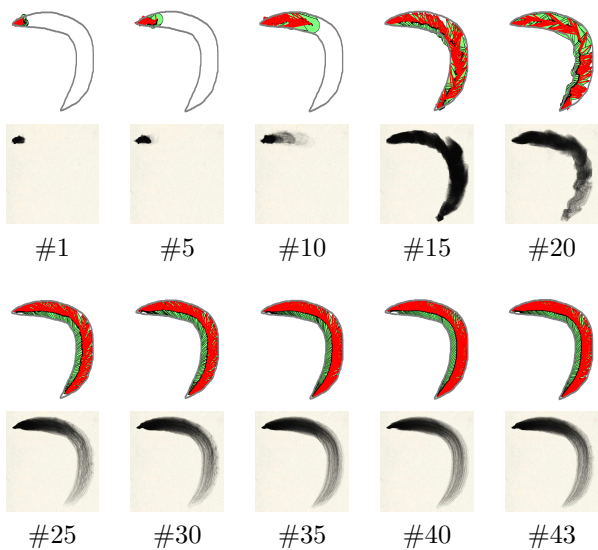

Figure 6. Examples of brush strokes in the learning process. The first row shows the paths of the footprints. The second row shows the rendering results. Over the iterations, the results become better.

each trial is computed over 300 training episode samples.

### 4.2. Results

The average return along the policy iteration is shown in Figure 5. The graph shows that the average return sharpy increases in an early stage and then it keeps stable after 35 iterations. Figure 6 shows examples of rendering and brush trajectory results in the policy training process.

Table 1 shows the performance of policies learned by our RL method and the DP method (Xie et al., 2011) on an Intel Core i7 2.70 GHz. According to the discussion in Xie et al. (2011), the performance of the DP method depends on the setup of the parameter in the energy function and the number of candidates in the DP search space. However, it is hard to manually find the optimal parameters in practice. In Table 1, we list results obtained by the DP method with changing the number of candidates in each step of the DP search space. The results of the expected return and the execution time are significantly different depending on the number of candidates. In the DP method, the best value of the return is 26.27 when the number of candidates is set to 180, but this is computationally very expensive ($2.08 \times 10^3$ seconds). Our RL method outperforms the best DP setup, with much less computation time.

Table 1. Comparison of the average return and the execution time between RL and DP.

| METHOD | # CANDIDATES | RETURN | TIME [SEC.] |
|---|---|---|---|
| DP | 5 | −0.60 | $\mathbf{3.95 \times 10^1}$ |
|  | 10 | −0.10 | $1.01 \times 10^2$ |
|  | 20 | 6.54 | $2.10 \times 10^2$ |
|  | 30 | 12.17 | $3.25 \times 10^2$ |
|  | 40 | 20.03 | $4.49 \times 10^2$ |
|  | 50 | 20.66 | $5.73 \times 10^2$ |
|  | 60 | 22.35 | $6.27 \times 10^2$ |
|  | 70 | 22.33 | $7.48 \times 10^2$ |
|  | 80 | 24.42 | $8.58 \times 10^2$ |
|  | 90 | 25.48 | $9.74 \times 10^2$ |
|  | 100 | 25.08 | $1.08 \times 10^3$ |
|  | 110 | 25.80 | $1.19 \times 10^3$ |
|  | 120 | 25.22 | $1.30 \times 10^3$ |
|  | 130 | 25.43 | $1.40 \times 10^3$ |
|  | 140 | 26.01 | $1.47 \times 10^3$ |
|  | 150 | 24.50 | $1.68 \times 10^3$ |
|  | 160 | 25.49 | $1.90 \times 10^3$ |
|  | 170 | 25.89 | $2.03 \times 10^3$ |
|  | 180 | **26.27** | $2.08 \times 10^3$ |
|  | 190 | 26.04 | $2.30 \times 10^3$ |
|  | 200 | 24.11 | $2.30 \times 10^3$ |
| RL | ∅ | **26.44** | $\mathbf{4.00 \times 10^1}$ |

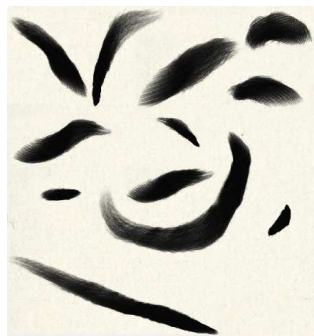

Figure 7. Results on new shapes.

We further apply our trained policy to more realistic shapes shown in Figure 7, which were not included in the train-ing samples. We can observe that the well-trained policy can produce smooth and natural brush strokes in various shapes. We therefore conclude that our RL method is useful in a practical environment.

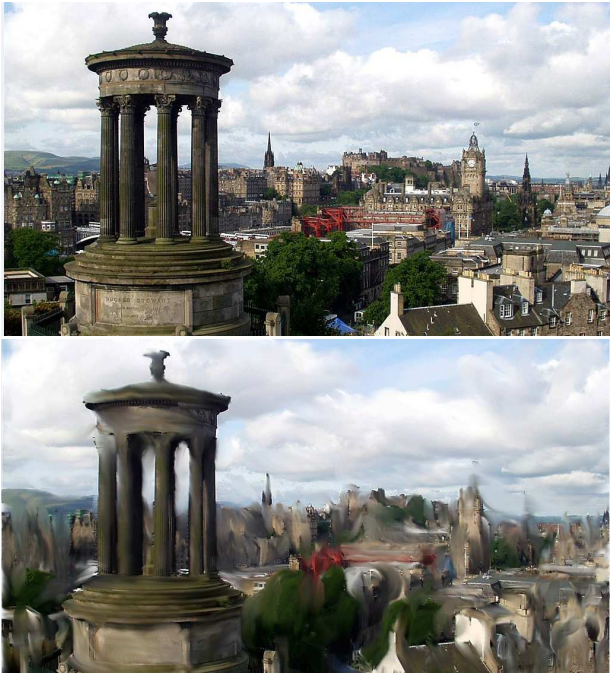

*Figure 8.* View of Edinburgh from Calton Hill. Upper: Real photo. Lower: Rendered result.

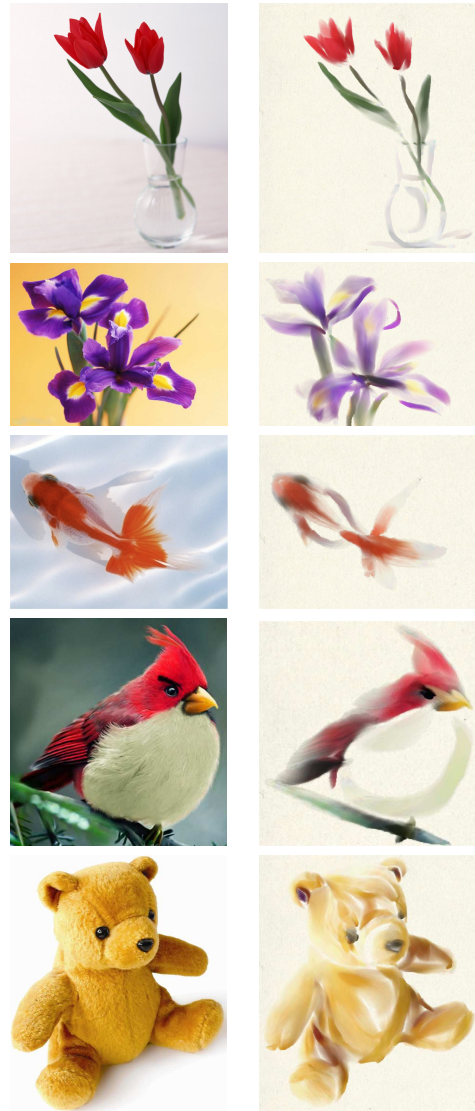

*Figure 9.* Results of automatic photo conversion into an oriental ink style. Left: Real photos. Right: Rendered results.

Finally, we apply our brush agent to automatic photo conversion into an oriental ink style. We manually drew contours on a photo and let the agent automatically fill the shapes with strokes. The results are shown in Figure 8 and Figure 9, which we think are of good quality.

## 5. Related Works

In this section, we briefly review current state-of-the-art in generating brush drawing, which have two approaches: *Physics-based painting* and *stroke-based rendering*.

### 5.1. Physics-Based Painting

This approach aims at reproducing a real painting process and giving users intuitive and natural feeling when holding a mouse or a pen-like device. Several previous works have dealt with modeling the brush shapes, its dynamics, and its interaction with the paper, and simulating the ink dispersion and absorption by the paper.

Among the first stream, early representative works include the hairy brushes (Strassmann, 1986) and the physics-based models (Saito & Nakajima, 1999; Chu & Tai, 2004; Chu et al., 2010). For interactive use, these virtual brushes are convenient to draw various styles of strokes. Despite the extensive research literature, controlling automatically a virtual brush with six degrees of freedom—three for the Cartesian coordinates and three for their angular orientation (pitch, roll, and yaw)—in addition to the dynamics of the tufts is complex and existing physics-based models are in fact simplifications of the real process.

On the other hand, while the digital painting tools provide expert users a professional environment with a canvas, brushes, mixing palettes, and a multitude of color options, non-expert users often prefer simplified environments where paintings can be generated with minimum interaction and painting expertise.

Another major problem is that the computational cost is usually very high for satisfactory visual effects to human eyes. Some of them rely on GPU acceleration for satisfactory speed performance (Chu et al., 2010). Also, due to over-simplification, none of these methods has been able to simulate certain special brush strokes such as those impasto ones created with paint knives.

### 5.2. Stroke-Based Rendering

In many situations, it is desirable to automatically convert real images into ink paintings, especially when the user has no painting expertise and is interested only in the painting results rather than in the painting process.

The skeleton stroke method (Hsu & Lee, 1994) generates brush strokes from the 2D paths given by either user interaction or automatic extraction from real images. However, the main difficulty is how to specify and vary the width of the strokes along the path as well as the texture of the strokes. One of the solutions is to specify the stroke backbone (Guo & Kunii, 2003) manually by a user. A limitation of such methods is that setting the values on each control point is time-consuming.

Contour-driven methods (Xie et al., 2011) can successfully generate strokes in desired shapes. However, there are several strict constraints: (I) When cutting the boundary region into slices at each step, the cross-sections should not intersect together. (II) A limited number of footprint candidates are only available for making the decision of moving to the next step. Although the second assumption ensures the same stride length of the agent at each step as well as speeding up the algorithm's execution time, states are typically modeled as discrete variables. This causes the resulting brush path not to be optimized well.

Our proposed approach belongs to the category of stroke-based rendering, with highly automatic and flexible stroke generation ability.

## 6. Conclusions

In this paper, we applied reinforcement learning to oriental ink painting, allowing automatic generation of smooth and natural strokes in arbitrary shapes. Our contributions include careful designs of actions, states, immediate rewards, and training sessions. One of the key ideas was to design the state space of the brush agent to be relative to its surrounding shape, which allows us to learn a general drawing policy independent of a specific entire shape. Another important idea was to train the brush agent locally in the given shape. This contributed highly to enhancing the generalization ability to new shapes, because even when a new shape is quite different from training data as a whole, it contains similar local shapes.

The experimental results demonstrated that our RL method gives better performance than the existing DP methods with much less computation time, and our RL agent can successfully draw new complex shapes with smooth and natural brush strokes. Also, applications to automatic photo conversion into an oriental style was demonstrated to be promising.

## Acknowledgments

The authors would like to thank to the anonymous reviewers for their helpful comments. Ning Xie was supported by MEXT Scholarship, Hirotaka Hachiya was supported by the FIRST program, and Masashi Sugiyama was supported by MEXT KAKENHI 23120004.